\documentclass[10pt,twocolumn,letterpaper]{article}

\usepackage[pagenumbers]{cvpr} % To force page numbers, e.g. for an arXiv version

\usepackage{graphicx}
\usepackage{amsmath}
\usepackage{amssymb}
\usepackage{booktabs}
\usepackage{threeparttable}
\usepackage[accsupp]{axessibility}

\usepackage[pagebackref,breaklinks,colorlinks]{hyperref}

\usepackage[capitalize]{cleveref}
\crefname{section}{Sec.}{Secs.}
\Crefname{section}{Section}{Sections}
\Crefname{table}{Table}{Tables}
\crefname{table}{Tab.}{Tabs.}

\def\cvprPaperID{7103} % *** Enter the CVPR Paper ID here
\def\confName{CVPR}
\def\confYear{2023}

\begin{document}

\title{DeSTSeg: Segmentation Guided Denoising Student-Teacher for Anomaly Detection}

\author{Xuan Zhang\textsuperscript{1}, \hspace{1em}Shiyu Li\textsuperscript{2}, \hspace{1em}Xi Li\textsuperscript{2}, \hspace{1em}Ping Huang\textsuperscript{2}, \hspace{1em}Jiulong Shan\textsuperscript{2}, \hspace{1em}Ting Chen\textsuperscript{1} \\
\textsuperscript{1}Tsinghua University\hspace{3em}\textsuperscript{2}Apple \\
{\tt\small x-zhang18@mails.tsinghua.edu.cn, \hspace{1em} \{shiyu\_li, weston\_li, huang\_ping, jlshan\}@apple.com,} \\
{\tt\small tingchen@tsinghua.edu.cn}
}

\maketitle

%%%%%%%%% ABSTRACT
\begin{abstract}

Visual anomaly detection, an important problem in computer vision, is usually formulated as a one-class classification and segmentation task. The student-teacher (S-T) framework has proved to be effective in solving this challenge. However, previous works based on S-T only empirically applied constraints on normal data and fused multi-level information. In this study, we propose an improved model called DeSTSeg, which integrates a pre-trained teacher network, a denoising student encoder-decoder, and a segmentation network into one framework. First, to strengthen the constraints on anomalous data, we introduce a denoising procedure that allows the student network to learn more robust representations. From synthetically corrupted normal images, we train the student network to match the teacher network feature of the same images without corruption. Second, to fuse the multi-level S-T features adaptively, we train a segmentation network with rich supervision from synthetic anomaly masks, achieving a substantial performance improvement. Experiments on the industrial inspection benchmark dataset demonstrate that our method achieves state-of-the-art performance, 98.6\% on image-level AUC, 75.8\% on pixel-level average precision, and 76.4\% on instance-level average precision.
\end{abstract}

%%%%%%%%% BODY TEXT

\section{Introduction}
\label{sec:intro}

\begin{figure*}[t]
  \centering
   \includegraphics[width=0.85\linewidth]{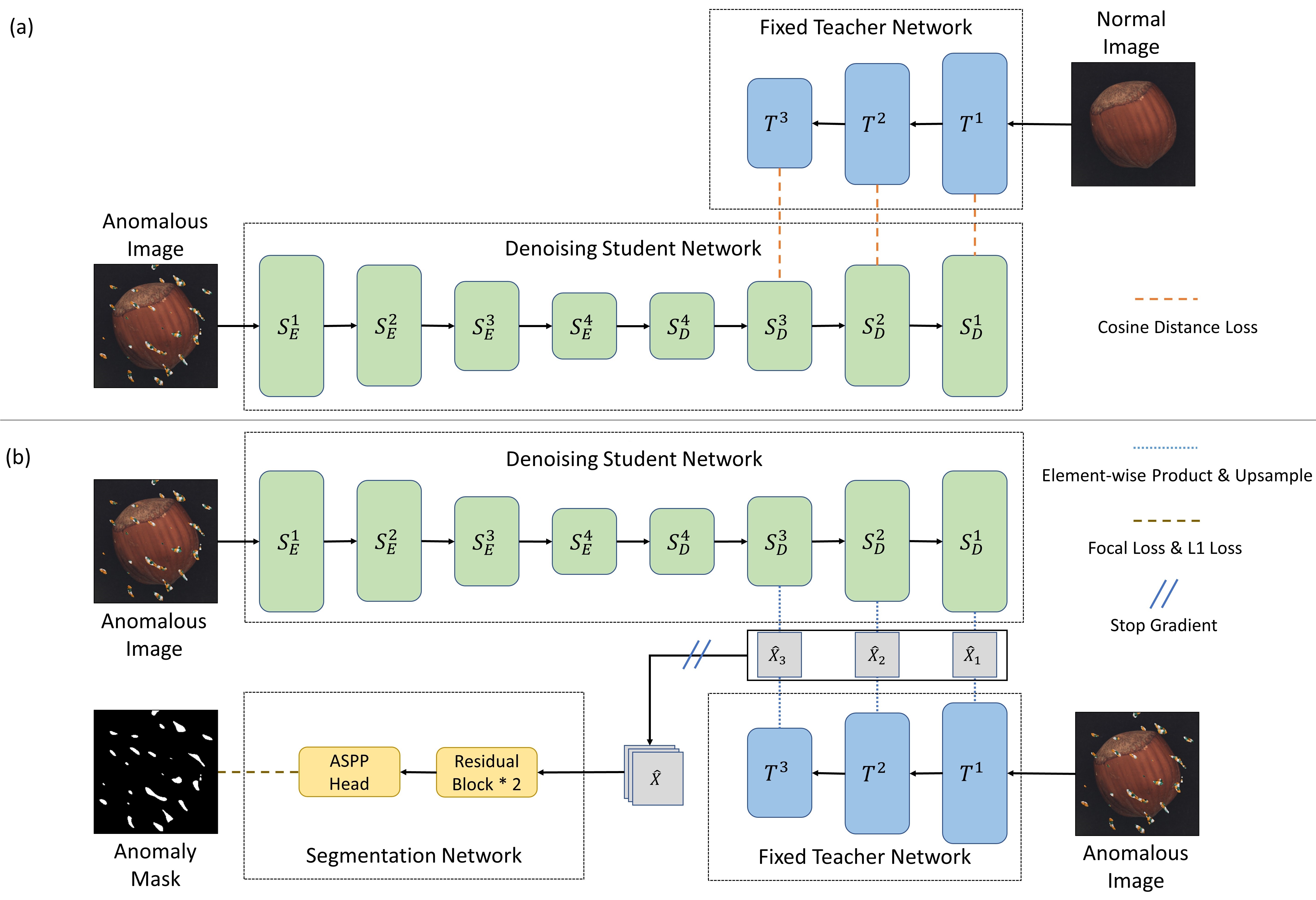}

   \caption{Overview of DeSTSeg. Synthetic anomalous images are generated and used during training. In the first step (a), the student network with synthetic input is trained to generate similar feature representations as the teacher network from the clean image. In the second step (b), the element-wise product of the student and teacher networks' normalized outputs are concatenated and utilized to train the segmentation network. The segmentation output is the predicted anomaly score map.}
   \label{fig:overview}
\end{figure*}

Visual anomaly detection (AD) with localization is an essential task in many computer vision applications such as industrial inspection ~\cite{roth2022towards, zavrtanik2021draem}, medical disease screening ~\cite{schlegl2019f, watanabe2019bone}, and video surveillance ~\cite{liu2018classifier, pang2020self}. The objective of these tasks is to identify both corrupted images and anomalous pixels in corrupted images. As anomalous samples occur rarely, and the number of anomaly types is enormous, it is unlikely to acquire enough anomalous samples with all possible anomaly types for training. Therefore, AD tasks were usually formulated as a one-class classification and segmentation, using only normal data for model training.

The student-teacher (S-T) framework, known as knowledge distillation, has proven effective in AD ~\cite{bergmann2020uninformed, wang2021student, deng2022anomaly, salehi2021multiresolution, yamada2021reconstruction}. In this framework, a teacher network is pre-trained on a large-scale dataset, such as ImageNet ~\cite{deng2009imagenet}, and a student network is trained to mimic the feature representations of the teacher network on an AD dataset with normal samples only. The primary hypothesis is that the student network will generate different feature representations from the teacher network on anomalous samples that have never been encountered in training. Consequently, anomalous pixels and images can be recognized in the inference phase. Notably, ~\cite{salehi2021multiresolution, wang2021student} applied knowledge distillation at various levels of the feature pyramid so that discrepancies from multiple layers were aggregated and demonstrated good performance. However, there is no guarantee that the features of anomalous samples are always different between S-T networks because there is no constraint from anomalous samples during the training. Even with anomalies, the student network may be over-generalized~\cite{pirnay2022inpainting} and output similar feature representations as those by the teacher network. Furthermore, aggregating discrepancies from multi-level in an empirical way, such as sum or product, could be suboptimal. For instance, in the MVTec AD dataset under the same context of ~\cite{wang2021student}, we observe that for the category of \textit{transistor}, employing the representation from the last layer, with 88.4\% on pixel-level AUC, outperforms that from the multi-level features, with 81.9\% on pixel-level AUC.

To address the problem mentioned above, we propose \textbf{DeSTSeg}, illustrated in \cref{fig:overview}, which consists of a denoising student network, a teacher network, and a segmentation network. We introduce random synthetic anomalies into the normal images and then use these corrupted images\footnote{All samples shown in this paper are licensed under the CC BY-NC-SA 4.0.} for training. The denoising student network takes a corrupted image as input, whereas the teacher network takes the original clean image as input. During training, the feature discrepancy between the two networks is minimized. In other words, the student network is trained to perform denoising in the feature space. Given anomalous images as input to both networks, the teacher network encodes anomalies naturally into features, while the trained denoising student network filters anomalies out of feature space. Therefore, the two networks are reinforced to generate distinct features from anomalous inputs. For the architecture of the denoising student network, we decided to use an encoder-decoder network for better feature denoising instead of adopting an identical architecture as the teacher network. In addition, instead of using empirical aggregation, we append a segmentation network to fuse the multi-level feature discrepancies in a trainable manner, using the generated binary anomaly mask as the supervision signal.

We evaluate our method on a benchmark dataset for surface anomaly detection and localization, MVTec AD~\cite{bergmann2019mvtec}. Extensive experimental results show that our method outperforms the state-of-the-art methods on image-level, pixel-level, and instance-level anomaly detection tasks. We also conduct ablation studies to validate the effectiveness of our proposed components. 

Our main contributions are summarized as follows. (1) We propose a denoising student encoder-decoder, which is trained to explicitly generate different feature representations from the teacher with anomalous inputs. (2) We employ a segmentation network to adaptively fuse the multi-level feature similarities to replace the empirical inference approach. (3) We conduct extensive experiments on the benchmark dataset to demonstrate the effectiveness of our method for various tasks. 

\section{Related Works}
\label{sec:related}

Anomaly detection and localization have been studied from numerous perspectives. In \textbf{image reconstruction}, researchers used autoencoder~\cite{bergmann2018improving}, variational autoencoder~\cite{vasilev2020q, baur2018deep} or generative adversarial network~\cite{schlegl2019f, schlegl2017unsupervised, perera2019ocgan} to train an image reconstruction model on normal data. The presumption is that anomalous images cannot be reconstructed effectively since they are not seen during training, so the difference between the input and reconstructed images can be used as pixel-level anomaly scores. However, anomaly regions still have a chance to be accurately reconstructed due to the over-generalization issue ~\cite{pirnay2022inpainting}. Another perspective is the \textbf{parametric density estimation}, which assumes that the extracted features of normal data obey a certain distribution, such as a multivariate Gaussian distribution~\cite{li2021cutpaste, defard2021padim, rippel2021modeling, lee2018simple}, and uses the normal dataset to estimate the parameters. Then, the outlier data are recognized as anomalous data by inference. Since the assumption of Gaussian distribution is too strict, some recent works borrow ideas from normalizing flow, by projecting an arbitrary distribution to a Gaussian distribution to approximate the density of any distributions~\cite{gudovskiy2022cflow, yu2021fastflow}. Besides, the \textbf{memory-based} approaches~\cite{roth2022towards, cohen2020sub, nazare2018pre, yi2020patch} build a memory bank of normal data in training. During inference, given a query item, the model selects the nearest item in the memory bank and uses the similarity between the query item and the nearest item to compute the anomaly score.

\textbf{Knowledge distillation. } Knowledge distillation is based on a pretrained teacher network and a trainable student network. As the student network is trained on an anomaly-free dataset, its feature representation of anomalies is expected to be distinct from that of the teacher network. Numerous solutions have been presented in the past to improve discrimination against various types of anomalies. For example, ~\cite{bergmann2020uninformed} used ensemble learning to train multiple student networks and exploited the irregularity of their feature representations to recognize the anomaly. ~\cite{wang2021student}, and ~\cite{salehi2021multiresolution} adopted multi-level feature representation alignment to capture both low-level and high-level anomalies. ~\cite{deng2022anomaly}, and ~\cite{yamada2021reconstruction} designed decoder architectures for the student network to avoid the shortcomings of identical architecture and the same dataflow between S-T networks. These works focus on improving the similarity of S-T representations on normal inputs, whereas our work additionally attempts to differentiate their representations on anomalous input.

\textbf{Anomaly simulation. } Although there is no anomalous data for training in the context of one-class classification AD, the pseudo-anomalous data could be simulated so that an AD model can be trained in a supervised way. Classical anomaly simulation strategies, such as rotation ~\cite{gidaris2018unsupervised} and cutout ~\cite{devries2017improved}, do not perform well in detecting fine-grained anomalous patterns~\cite{li2021cutpaste}. A simple yet effective strategy is called CutPaste ~\cite{li2021cutpaste} that randomly selects a rectangular region inside the original image and then copies and pastes the content to a different location within the image. Another strategy proposed in ~\cite{zavrtanik2021draem} and also adopted in ~\cite{zavrtanik2022dsr} used two-dimensional Perlin noise to simulate a more realistic anomalous image. With the simulated anomalous images and corresponding ground truth masks, ~\cite{zavrtanik2021draem, zavrtanik2022dsr, song2021anoseg} localized anomalies with segmentation networks. In our system, we adopt the ideas of ~\cite{zavrtanik2021draem} for anomaly simulation and segmentation.

\section{Method}
\label{sec:method}

The proposed DeSTSeg consists of three main components: a pre-trained teacher network, a denoising student network, and a segmentation network.

As illustrated in \cref{fig:overview}, synthetic anomalies are introduced into normal training images, and the model is trained in two steps. In the first step, the simulated anomalous image is utilized as the student network input, whereas the original clean image is the input to the teacher network. The weights of the teacher network are fixed, but the student network for denoising is trainable. In the second step, the student model is fixed as well. Both the student and teacher networks take the synthetic anomaly image as their input to optimize parameters in the segmentation network to localize the anomalous regions. For inference, pixel-level anomaly maps are generated in an end-to-end mode, and the corresponding image-level anomaly scores can be computed via post-processing.

\subsection{Synthetic Anomaly Generation}

The training of our model relies on synthetic anomalous images which are generated using the same algorithm proposed in ~\cite{zavrtanik2021draem}. Random two-dimensional Perlin noise is generated and binarized by a preset threshold to obtain an anomaly mask $M$. An anomalous image $I_a$ is generated by replacing the mask region with a linear combination of an anomaly-free image $I_n$ and an arbitrary image from external data source $A$, with an opacity factor $\beta$ randomly chosen between $[0.15, 1]$.

\begin{equation}
  I_a = \beta(M \odot A) + (1-\beta)(M \odot I_n) + (1-M)\odot I_n
  \label{eq:anomalysimulation}
\end{equation}

$\odot$ means the element-wise multiplication operation. The anomaly generation is performed online during training. By using this algorithm, three benefits are introduced. First, compared to painting a rectangle anomaly mask ~\cite{li2021cutpaste}, the anomaly mask generated by random Perlin noise is more irregular and similar to actual anomalous shapes. Second, the image used as anomaly content $A$ could be arbitrarily chosen without elaborate selection ~\cite{zavrtanik2021draem}. Third, the introduction of opacity factor $\beta$ can be regarded as a data augmentation~\cite{zhang2017mixup} to effectively increase the diversity of the training set.

\subsection{Denoising Student-Teacher Network}
In previous multi-level knowledge distillation approaches~\cite{wang2021student, salehi2021multiresolution}, the input of the student network (normal image) is identical to that of the teacher network, as is the architecture of the student network. However, our proposed denoising student network and the teacher network take paired anomalous and normal images as input, with the denoising student network having a distinct encoder-decoder architecture. In the following two paragraphs, we will examine the motivation for this design.

First, as mentioned in \cref{sec:intro}, an optimization target should be established to encourage the student network to generate anomaly-specific features that differ from the teacher’s. We further endow a more straightforward target to the student network: to build normal feature representations on anomalous regions supervised by the teacher network. As the teacher network has been pre-trained on a large dataset, it can generate discriminative feature representations in both normal and anomalous regions. Therefore, the denoising student network will generate different feature representations from those by the teacher network during inference. Besides, as mentioned in \cref{sec:related}, the memory-based approaches look for the most similar normal item in the memory bank to the query item and use their similarity for inference. Similarly, we optimize the denoising student network to reconstruct the normal features.

Second, given the feature reconstruction task, we conclude that the student network should not copy the architecture of the teacher network. Considering the process of reconstructing the feature of an early layer, it is well known that the lower layers of CNN capture local information, such as texture and color. In contrast, the upper layers of CNN express global semantic information~\cite{deng2022anomaly}. Recalling that our denoising student network should reconstruct the feature of the corresponding normal image from the teacher network, such a task relies on global semantic information of the image and could not be done perfectly with only a few lower layers. We notice that the proposed task design resembles image denoising, with the exception that we wish to denoise the image in the feature space. The encoder-decoder architecture is widely used for image denoising. Therefore, we adopted it as the denoising student network's architecture. There is an alternative way to use the teacher network as an encoder and reverse the student network as the decoder~\cite{deng2022anomaly, yamada2021reconstruction}; however, our preliminary experimental results show that a complete encoder-decoder student network performs better. One possible explanation is that the pre-trained teacher network is usually trained on ImageNet with classification tasks; thus, the encoded features in the last layers lack sufficient information to reconstruct the feature representations at all levels.

Following ~\cite{wang2021student}, the teacher network is an ImageNet pre-trained ResNet18~\cite{he2016deep} with the final block removed (i.e., conv5\_x). The output feature maps are extracted from the three remaining blocks, i.e., conv2\_x, conv3\_x, and conv4\_x denoted as $T^1$, $T^2$, and $T^3$, respectively. Regarding the denoising student network, the encoder is a randomly initialized ResNet18 with all blocks, named $S_{E}^1$, $S_{E}^2$, $S_{E}^3$, and $S_{E}^4$, respectively. The decoder is a reversed ResNet18 (by replacing all downsampling with bilinear upsampling) with four residual blocks, named $S_{D}^4$, $S_{D}^3$, $S_{D}^2$, and $S_{D}^1$, respectively.

We minimize the cosine distance between features from $T^k$ and $S_D^k$, $k=1,2,3$. Denoting $F_{T_k} \in \mathcal{R}^{C_k\times H_k\times W_k}$ the feature representation from layer $T^k$, and $F_{S_k} \in \mathcal{R}^{C_k\times H_k\times W_k}$ the feature representation from layer $S_{D}^k$, the cosine distances can be computed through ~\cref{eq:defX} and ~\cref{eq:defD}. $i$ and $j$ stand for the spatial coordinate on the feature map. In particular, $i=1...H_k$ and $j=1...W_k$. The loss is the sum of distances across three different feature levels as shown in ~\cref{eq:cosineloss}.

\begin{equation}
  X_k(i,j)=\frac{F_{T_k}(i,j)\odot F_{S_k}(i,j)}{||F_{T_k}(i,j)||_2||F_{S_k}(i,j)||_2}
  \label{eq:defX}
\end{equation}
\begin{equation}
  D_k(i,j)=1 - \sum_{c=1}^{C_k} X_k(i,j)_c
  \label{eq:defD}
\end{equation}
\begin{equation}
  L_{cos} = \sum_{k=1}^{3}\left(\frac{1}{H_k W_k}\sum_{i,j=1}^{H_k,W_k}D_k(i,j)\right)
  \label{eq:cosineloss}
\end{equation}

\begin{figure*}[t]
  \centering
   \includegraphics[width=0.8\linewidth]{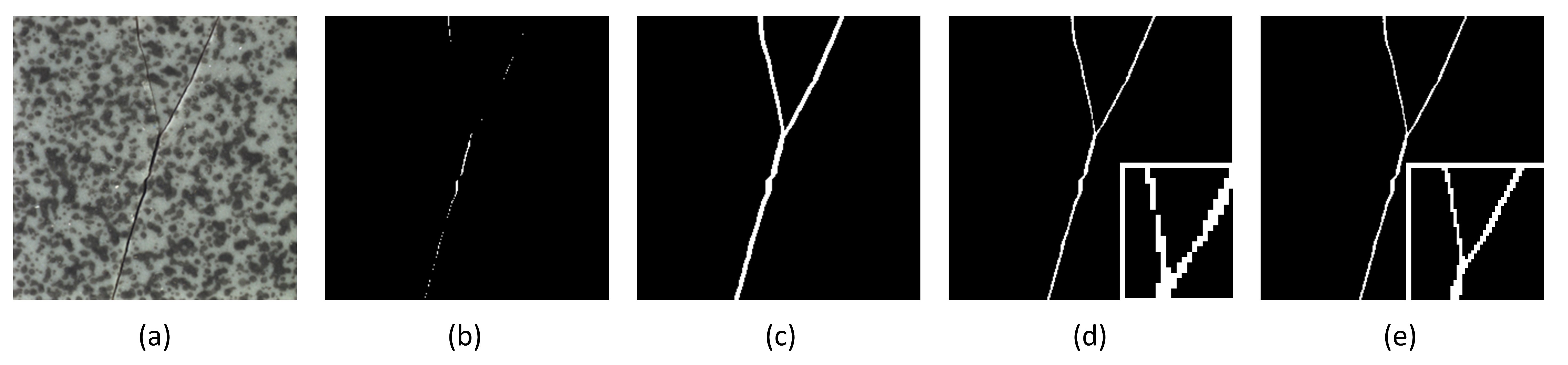}

   \caption{The binary ground truth is downsampled with different implementations. (a) The grid image with a crack anomaly. (b) Downsample with bilinear interpolation, then floor all values between $(0, 1)$ to zero\cite{roth2022towards, gudovskiy2022cflow}. The mask has almost vanished. (c) Downsample with bilinear interpolation, then ceil all values between $(0, 1)$ to one\cite{zavrtanik2021draem}. The mask is thicker than expected. (d) Downsample with the nearest interpolation, interrupting the original contiguous region. (e) Our proposed approach, downsample with bilinear interpolation and round values by threshold=0.5, The original contiguous region is not interrupted.}
   \label{fig:downsample}
\end{figure*}
\subsection{Segmentation Network}
\label{subsec:segmentation}

In ~\cite{wang2021student, salehi2021multiresolution}, the cosine distances from multi-level features are summed up directly to represent the anomaly score of each pixel. However, the results can be suboptimal if discriminations of all level features are not equally accurate. To address this issue, we add a segmentation network to guide the feature fusion with additional supervision signals.

We freeze the weights of both the student and teacher networks to train the segmentation network. The synthetic anomalous image is utilized as the input for both S-T networks, and the corresponding binary anomaly mask is the ground truth. The similarities of the feature maps $(T^1, S_D^1)$, $(T^2, S_D^2)$, $(T^3, S_D^3)$ are calculated by ~\cref{eq:defX} and upsampled to the same size as $X_1$, which is 1/4 of the input size. The upsampled features, denoted as $\hat{X}_1$, $\hat{X}_2$, and $\hat{X}_3$, are then concatenated as $\hat{X}$, which is fed into the segmentation network. We also investigate alternative ways to compute the input of the segmentation network in \cref{subsec:ablation}. The segmentation network contains two residual blocks and one Atrous Spatial Pyramid Pooling (ASPP) module~\cite{chen2017deeplab}. There is no upsampling or downsampling; thus, the output size equals the size of ${X}_1$. Although this may lead to resolution loss to some extent, it reduces the memory cost for training and inference, which is crucial in practice.

The segmentation training is optimized by employing the focal loss~\cite{lin2017focal} and the L1 loss. In the training set, the majority of pixels are normal and easily recognized as background. Only a small portion of the image consists of anomalous pixels that must be segmented. Therefore, the focal loss can help the model to focus on the minority category and difficult samples. In addition, the L1 loss is employed to improve the sparsity of the output so that the segmentation mask’s boundaries are more distinct. To compute the loss, we downsample the ground truth anomaly mask to a size equal to 1/4 of the input size, which matches the output $(H_1, W_1)$. Mathematically, we denote the output probability map as $\hat{Y}$ and the downsampled anomaly mask as $M$, and the focal loss is computed using \cref{eq:focal} where $p_{ij}=M_{ij}\hat{Y}_{ij} + (1 - M_{ij})(1 - \hat{Y}_{ij})$ and $\gamma$ is the focusing parameter. The L1 loss is computed by \cref{eq:l1}, and the segmentation loss is computed by \cref{eq:seg}.

\begin{equation}
  L_{focal} = -\frac{1}{H_1W_1}\sum_{i,j=1}^{H_1,W_1}(1-p_{ij})^{\gamma}\log(p_{ij})
  \label{eq:focal}
\end{equation}
\begin{equation}
  L_{l1} = \frac{1}{H_1W_1}\sum_{i,j=1}^{H_1,W_1}|M_{ij}-\hat{Y}_{ij}|
  \label{eq:l1}
\end{equation}
\begin{equation}
  L_{seg} = L_{focal}+L_{l1}
  \label{eq:seg}
\end{equation}

\subsection{Inference}
In the inference stage, the test image is fed into both the teacher and student networks. The segmentation prediction is finally upsampled to the input size and taken as the anomaly score map. It is expected that anomalous pixels in the input image will have greater values in the output. To calculate the image-level anomaly score, we use the average of the top $T$ values from the anomaly score map, where $T$ is a tuning hyperparameter.

\section{Experiments}
\label{sec:experiments}

\subsection{Dataset}
We evaluate our method using the MVTec AD ~\cite{bergmann2019mvtec} dataset, which is one of the most widely used benchmarks for anomaly detection and localization. The dataset comprises 15 categories, including 10 objects and 5 textures. For each category, there are hundreds of normal images for training and a mixture of anomalous and normal images for evaluation. The image sizes range from $700\times700$ to $1024\times1024$ pixels. For evaluation purposes, pixel-level binary annotations are provided for anomalous images in the test set. In addition, the Describable Textures Dataset (DTD)~\cite{6909856} is used as the anomaly source image $A$ in ~\cref{eq:anomalysimulation}. ~\cite{zavrtanik2021draem} showed that other datasets such as ImageNet can achieve comparable performance but DTD is much smaller and easy to use.

%% SHIYU
% The Describable Textures Dataset (DTD) [6] is used as the anomaly source image A in Eq. (1). [38] showed that other datasets such as ImageNet can achieve comparable performance but  DTD is much smaller and easy to use.

\begin{table*}[htbp]
  \centering
  \begin{threeparttable}
  \setlength{\tabcolsep}{3mm}{
  \begin{tabular}{@{}llllllll}
    \toprule
    & US~\cite{bergmann2020uninformed} & STPM~\cite{wang2021student} & CutPaste~\cite{li2021cutpaste} & DRAEM~\cite{zavrtanik2021draem} & DSR~\cite{zavrtanik2022dsr} & PatchCore~\cite{roth2022towards} & Ours \\
    \midrule
    & 87.7 & 95.1 & 95.2 & 98.0 & 98.2 & 98.5 & \pmb{98.6}{\scriptsize$\pm$0.4} \\
    \bottomrule
  \end{tabular}}
  \end{threeparttable}

  % \caption{Image-level anomaly detection results on MVTec AD dataset (AUC \%). Results are averaged over all categories.}
  \caption{Image-level anomaly detection AUC (\%) on MVTec AD dataset. Results are averaged over all categories.}
  \label{tab:anomalydetection}

\end{table*}

\begin{table*}[htbp]
  \centering
  \begin{threeparttable}
  \setlength{\tabcolsep}{1.5mm}{
  \begin{tabular}{@{}llllllll}
    \toprule
    & US~\cite{bergmann2020uninformed} & STPM~\cite{wang2021student} & CutPaste~\cite{li2021cutpaste} & DRAEM~\cite{zavrtanik2021draem} & DSR~\cite{zavrtanik2022dsr} & PatchCore~\cite{roth2022towards} & Ours \\
    \midrule
    bottle & 97.8 / 74.2 & 98.8 / 80.6 & 97.6 / - & \pmb{99.3} / 89.8 & - / \pmb{91.5} & 98.9 / 80.1 & 99.2{\scriptsize$\pm$0.2} / 90.3{\scriptsize$\pm$1.8} \\
    cable & 91.9 / 48.2 & 94.8 / 58.0 & 90.0 / - & 95.4 / 62.6 & - / \pmb{70.4} & \pmb{98.8} / 70.0 & 97.3{\scriptsize$\pm$0.4} / 60.4{\scriptsize$\pm$2.3} \\
    capsule & 96.8 / 25.9 & 98.2 / 35.9 & 97.4 / - & 94.1 / 43.5 & - / 53.3 & \pmb{99.1} / 48.1 & \pmb{99.1}{\scriptsize$\pm$0.0} / \pmb{56.3}{\scriptsize$\pm$1.1} \\
    carpet & 93.5 / 52.2 & \pmb{99.1} / 65.3 & 98.3 / - & 96.2 / 64.4 & - / \pmb{78.2} & \pmb{99.1} / 66.7 & 96.1{\scriptsize$\pm$2.2} / 72.8{\scriptsize$\pm$5.8} \\
    grid & 89.9 / 10.1 & 99.1 / 45.4 & 97.5 / - & \pmb{99.5} / 56.8 & - / \pmb{68.0} & 98.9 / 41.0 & 99.1{\scriptsize$\pm$0.1} / 61.5{\scriptsize$\pm$1.6} \\
    hazelnut & 98.2 / 57.8 & 98.9 / 60.3 & 97.3 / - & 99.5 / 88.1 & - / 87.3 & 99.0 / 61.5 & \pmb{99.6}{\scriptsize$\pm$0.2} / \pmb{88.4}{\scriptsize$\pm$2.2} \\
    leather & 97.8 / 40.9 & 99.2 / 42.9 & 99.5 / - & 98.9 / 69.9 & - / 62.5 & 99.4 / 51.0 & \pmb{99.7}{\scriptsize$\pm$0.0} / \pmb{75.6}{\scriptsize$\pm$1.2} \\
    metal\_nut & 97.2 / 83.5 & 97.2 / 79.3 & 93.1 / - & 98.7 / 91.7 & - / 67.5 & \pmb{98.8} / 88.8 & 98.6{\scriptsize$\pm$0.4} / \pmb{93.5}{\scriptsize$\pm$1.1} \\
    pill & 96.5 / 62.0 & 94.7 / 63.3 & 95.7 / - & 97.6 / 46.1 & - / 65.7 & 98.2 / 78.7 & \pmb{98.7}{\scriptsize$\pm$0.4} / \pmb{83.1}{\scriptsize$\pm$4.2} \\
    screw & 97.4 / 7.8 & 98.6 / 26.9 & 96.7 / - & \pmb{99.7} / \pmb{71.5} & - / 52.5 & 99.5 / 41.4 & 98.5{\scriptsize$\pm$0.3} / 58.7{\scriptsize$\pm$3.7} \\
    tile & 92.5 / 65.3 & 96.6 / 61.7 & 90.5 / - & \pmb{99.5} / \pmb{96.9} & - / 93.9 & 96.6 / 59.3 & 98.0{\scriptsize$\pm$0.7} / 90.0{\scriptsize$\pm$2.5} \\
    toothbrush & 97.9 / 37.7 & 98.9 / 48.8 & 98.1 / - & 98.1 / 54.7 & - / 74.2 & 98.9 / 51.6 & \pmb{99.3}{\scriptsize$\pm$0.1} / \pmb{75.2}{\scriptsize$\pm$1.8} \\
    transistor & 73.7 / 27.1 & 81.9 / 44.4 & 93.0 / - & 90.0 / 51.7 & - / 41.1 & \pmb{96.2} / 63.2 & 89.1{\scriptsize$\pm$3.4} / \pmb{64.8}{\scriptsize$\pm$4.0} \\
    wood & 92.1 / 53.3 & 95.2 / 47.0 & 95.5 / - & 97.0 / 80.5 & - / 68.4 & 95.1 / 52.3 & \pmb{97.7}{\scriptsize$\pm$0.3} / \pmb{81.9}{\scriptsize$\pm$1.2} \\
    zipper & 95.6 / 36.1 & 98.0 / 54.9 & 99.3 / - & 98.6 / 72.3 & - / 78.5 & 99.0 / 64.0 & \pmb{99.1}{\scriptsize$\pm$0.5} / \pmb{85.2}{\scriptsize$\pm$3.3} \\
    \midrule
    average & 93.9 / 45.5 & 96.6 / 54.3 & 96.0 / - & 97.5 / 69.3 & - / 70.2 & \pmb{98.4} / 61.2 & 97.9{\scriptsize$\pm$0.3} / \pmb{75.8}{\scriptsize$\pm$0.8} \\
    \bottomrule
  \end{tabular}}
  \end{threeparttable}

  % \caption{Pixel-level anomaly localization results on MVTec AD Dataset (AUC/AP \%).}
  \caption{Pixel-level anomaly localization AUC / AP (\%) on MVTec AD dataset.}
  \label{tab:anomalylocalization}

\end{table*}

\subsection{Evaluation Metrics}
\label{subsec:metrics}

\textbf{Image-level evaluation.} Following the previous work in anomaly detection work, AUC (i.e., area under the ROC curve) is utilized to evaluate image-level anomaly detection.

\textbf{Pixel-level evaluation.} AUC is also selected to evaluate the pixel-level result. Additionally, we report average precision (AP) since it is a more appropriate metric for heavily imbalanced classes~\cite{saito2015precision}.

\textbf{Instance-level evaluation.} In real-world applications, such as industrial defect inspection and medical imaging lesion detection, users are more concerned with whether the model can fully or partially localize an instance than with each individual pixel. In ~\cite{bergmann2020uninformed}, per-region-overlap (PRO) is proposed, which equally weights the connected components of different sizes in the ground truth. It computes the overlap between prediction and ground truth within a user-specified false positive rate (30\%). However, because instance recall is essential in practice, we propose to use instance average precision (IAP) as a more straightforward metric. Formally, we define an anomaly instance as a maximally connected ground truth region. Given a prediction map, an anomalous instance is considered detected if and only if more than 50\% of the region pixels are predicted as positive. Under different thresholds, a list of pixel-level precision and instance-level recall rate points can be drawn as a curve. The average precision of this curve is calculated as IAP. For those applications requiring an extremely high recall, the precision at $recall=k\%$ is also computed and denoted as IAP@k. In our experiments, we evaluate our model under a high-stakes scenario by setting $k=90$. 

\textbf{Ground truth downsampling method.} We notice that the prior implementations of pixel-level evaluation are poorly aligned. Most of the works downsampled the ground truth to $256\times256$ for faster computation, but some performed an extra $224\times224$ center crop ~\cite{roth2022towards, defard2021padim, cohen2020sub}. In addition, the downsampling implementations are not standardized either ~\cite{zavrtanik2021draem, roth2022towards, gudovskiy2022cflow}, resulting in the varying ground truth and unfair evaluation. In some cases, the downsampling introduces severe distortion, as illustrated in \cref{fig:downsample}. In our work, we use bilinear interpolation to downsample the binary mask to $256\times256$ and then round the result with a threshold of 0.5. This implementation can preserve the continuity of the original ground truth mask without over or under-estimating.

\subsection{Results}

In order to make fair comparisons with other works, we re-evaluated the official pre-trained models of \cite{wang2021student}, \cite{roth2022towards}, and \cite{zavrtanik2021draem} using our proposed evaluation introduced in \cref{subsec:metrics}. For methods without open-source code, we use the results mentioned in the original papers. Unavailable results are denoted with `-'. We repeat the experiments of our method 5 times with different random seeds to report the standard deviation.

% SHIYU
% All the compared methods apply models with similar size as ours, and we do not consider methods that use very large models like WideResNet-101\cite{zagoruyko2016wide} and Transformer\cite{vaswani2017attention} for fairness 
% 这句话不要了吧

{\bf Image-level anomaly detection. }We report the AUC for the image-level anomaly detection task in \cref{tab:anomalydetection}. The performance of our method reaches state-of-the-art on average. Category-specific results are shown in the supplementary material.

{\bf Pixel-level anomaly localization. }We report the AUC and AP values for the pixel-level anomaly localization task in \cref{tab:anomalylocalization}. On average, our method outperforms state-of-the-art by 5.6\% on AP and achieves AUC scores comparable to PatchCore~\cite{roth2022towards}. Our method reaches the highest or near-highest score in the majority of categories, indicating that our approach generalizes well over a wide range of industrial application scenes.

\begin{table*}[htbp]
  \centering
  \begin{threeparttable}
  \setlength{\tabcolsep}{5mm}{
  \begin{tabular}{@{}lllll}
    \toprule
    & STPM~\cite{wang2021student} & DRAEM~\cite{zavrtanik2021draem} & PatchCore~\cite{roth2022towards} & Ours \\
    \midrule
    bottle & 83.2 / 73.3 & 90.3 / \pmb{84.8} & 81.8 / 70.1 & \pmb{90.5}{\scriptsize$\pm$1.7} / 82.5{\scriptsize$\pm$4.1} \\
    cable & 54.9 / 17.2 & 47.0 / 10.8 & \pmb{69.2} / \pmb{50.6} & 51.1{\scriptsize$\pm$2.5} / 26.7{\scriptsize$\pm$3.7} \\
    capsule & 37.2 / 17.9 & \pmb{50.7} / 21.4 & 44.2 / 26.9 & 49.4{\scriptsize$\pm$1.5} / \pmb{27.3}{\scriptsize$\pm$3.3} \\
    carpet & 68.4 / 52.2 & 76.8 / 32.3 & 64.4 / 43.7 & \pmb{84.5}{\scriptsize$\pm$4.9} / \pmb{58.6}{\scriptsize$\pm$17.1} \\
    grid & 45.7 / 21.0 & 55.5 / 42.3 & 39.1 / 15.6 & \pmb{61.6}{\scriptsize$\pm$1.8} / \pmb{47.4}{\scriptsize$\pm$2.9} \\
    hazelnut & 64.8 / 56.2 & \pmb{95.7} / \pmb{89.0} & 63.8 / 52.5 & 87.7{\scriptsize$\pm$1.8} / 77.6{\scriptsize$\pm$3.4} \\
    leather & 46.2 / 24.9 & \pmb{78.6} / 55.0 & 50.1 / 30.1 & 77.5{\scriptsize$\pm$1.8} / \pmb{65.3}{\scriptsize$\pm$3.9} \\
    metal\_nut & 83.4 / 81.7 & 92.6 / 83.9 & 90.1 / 84.6 & \pmb{93.6}{\scriptsize$\pm$1.3} / \pmb{86.5}{\scriptsize$\pm$2.7} \\
    pill & 72.0 / 45.5 & 46.9 / 41.5 & 82.7 / \pmb{63.5} & \pmb{84.8}{\scriptsize$\pm$3.8} / 61.1{\scriptsize$\pm$12.4} \\
    screw & 24.4 / 4.2 & \pmb{68.8} / \pmb{33.0} & 38.4 / 16.3 & 53.6{\scriptsize$\pm$3.6} / 8.6{\scriptsize$\pm$2.3} \\
    tile & 62.9 / 55.3 & \pmb{98.9} / \pmb{98.2} & 60.0 / 52.1 & 94.7{\scriptsize$\pm$1.8} / 86.5{\scriptsize$\pm$3.6} \\
    toothbrush & 41.9 / 23.4 & 44.7 / 21.5 & 40.4 / 22.1 & \pmb{59.8}{\scriptsize$\pm$2.9} / \pmb{32.1}{\scriptsize$\pm$5.1} \\
    transistor & 53.4 / 8.5 & 59.3 / 22.8 & 69.9 / 36.8 & \pmb{78.3}{\scriptsize$\pm$2.5} / \pmb{49.6}{\scriptsize$\pm$8.4} \\
    wood & 56.0 / 35.4 & \pmb{88.4} / 72.6 & 59.7 / 35.6 & 87.8{\scriptsize$\pm$2.8} / \pmb{76.4}{\scriptsize$\pm$3.4} \\
    zipper & 59.1 / 46.6 & 78.7 / 67.0 & 66.0 / 52.4 & \pmb{90.6}{\scriptsize$\pm$2.3} / \pmb{80.3}{\scriptsize$\pm$4.9} \\
    \midrule
    average & 56.9 / 37.5 & 71.5 / 51.7 & 61.3 / 43.5 & \pmb{76.4}{\scriptsize$\pm$1.0} / \pmb{57.8}{\scriptsize$\pm$1.8} \\
    \bottomrule
  \end{tabular}}
  \end{threeparttable}

  % \caption{Instance-level anomaly detection results on MVTec AD Dataset (IAP/IAP@90 \%).}
  \caption{Instance-level anomaly detection IAP / IAP@90 (\%) on MVTec AD dataset.}
  \label{tab:instanceanomaly}

\end{table*}

{\bf Instance-level anomaly detection.} The IAP and IAP@90 of the instance-level anomaly detection are reported in \cref{tab:instanceanomaly}. Our method achieves the state of the art for both metrics. On average, our approach reaches an IAP@90 of 57.8\%, which indicates that when 90\% of anomaly instances are detected, the pixel-level precision is 57.8\%, or equally, the pixel-level false positive rate is 42.2\%. As some categories (e.g., carpet, pill) contain hard samples close to the decision boundary, their standard deviations of IAP@90 are relatively high. In practice, these metrics can be used to determine whether the performance is acceptable for an application.

\begin{figure}[t]
  \centering
   \includegraphics[width=\linewidth]{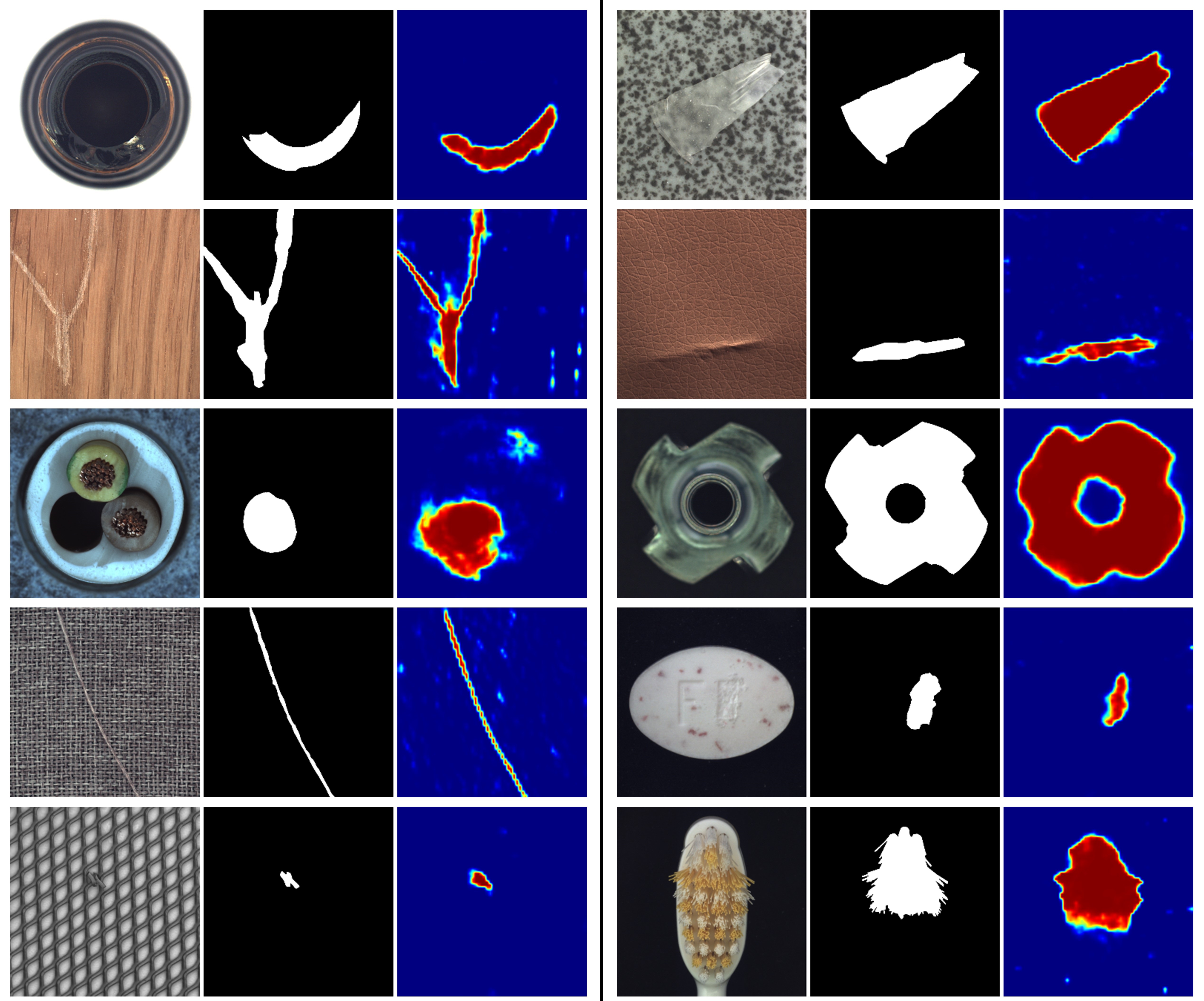}

   \caption{Visualization examples of our method. For each example, left: input image; middle: ground truth; right: prediction map.}
   \label{fig:visualexample}
\end{figure}

\begin{figure}[t]
  \centering
   \includegraphics[width=0.8\linewidth]{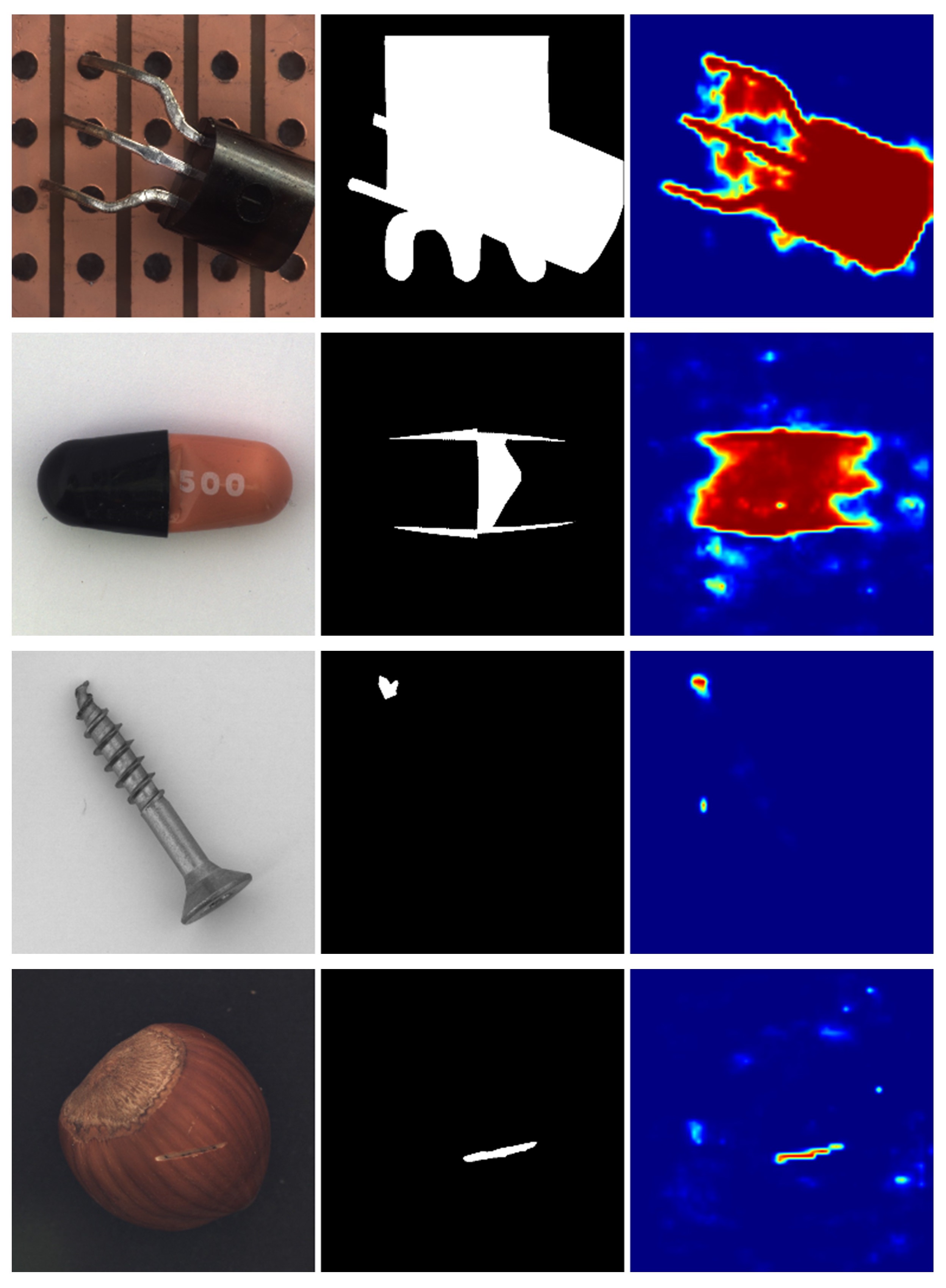}

   \caption{Failure cases of our method. The examples are chosen from transistor, capsule, screw, and hazelnut (from top to bottom). For each example, left: input image; middle: ground truth; right: prediction map.}
   \label{fig:failureexample}
\end{figure}

{\bf Category-specific analysis.} For the category cable, memory-based approaches ~\cite{roth2022towards, zavrtanik2022dsr} have better performance than ours since the normal pixels have larger intra-class distances than categories with periodic textures. For the categories grid, screw, and tile, the anomalies are relatively small or thin. Therefore, methods with higher resolution predictions, such as ~\cite{zavrtanik2021draem, zavrtanik2022dsr}, can achieve higher performance, but require more memory and computation. For the remaining categories, our method achieves comparable or higher performance than the compared methods.

{\bf Visualization examples.} Several visualization examples of our method from various categories are presented in \cref{fig:visualexample}. Our method can precisely localize the anomaly regions. More examples are shown in the supplementary material.

{\bf Analysis of failure cases.} We analyze some failure cases illustrated in \cref{fig:failureexample}. On the one hand, several ambiguous ground truths are responsible for a number of failure occurrences. In a transistor case from the first row, the ground truth highlights both the original and misplaced location, while the prediction mask only covers the misplaced location. For a capsule case shown in the second tow, the ground truth contains most of the distorted parts, whereas the prediction mask covers the entire capsule. In these cases, we would argue that our predictions are still useful.

On the other hand, some failure cases, such as those shown in the third and fourth rows, result from noisy backgrounds. Tiny fibers and stains are highlighted due to the susceptibility of our model. We leave it to future work to investigate whether these anomalies are acceptable in order to draw more accurate conclusions.

\subsection{Ablation Studies}
\label{subsec:ablation}

\textbf{Network architecture.} In \cref{tab:mainablation}, we evaluate the effectiveness of our three designs: replacing the training inputs of the vanilla student network with synthetic anomalies to enable a denoising procedure (\textbf{den}), applying encoder-decoder architecture to the student network(\textbf{ed}), and appending the segmentation network (\textbf{seg}) to replace the empirical feature fusion strategy, i.e., a product of cosine distances\cite{wang2021student}. (a) Comparing experiments 1 and 2, it can be found that only changing the student network's input to anomalous images undermines performance. However, experiment 5 shows improvement when \textbf{ed} is added, indicating that the \textbf{den} can be boosted by adopting \textbf{ed} architecture. (b) The comparisons of experiments 1 with 4, 2 with 6, 3 with 7, and 5 with 8, showcase that the segmentation network can significantly improve the performances of all three metrics. (c) Comparing experiments 4 and 8, it can be found that the combination of \textbf{den} and \textbf{ed} provides more useful features for the segmentation network than a vanilla S-T network does. The best result is achieved by combining all three main designs.

% SHIYU
% The comparisons between experiments 1 and 4, 2 and 6, 3 and 7, 5 and 8, showcase that

\textbf{Segmentation loss.} In \cref{tab:otherablation}, we examine the effectiveness of the L1-loss in the segmentation loss (\cref{eq:seg}). It can be observed that the L1-loss improves performance.

\textbf{Segmentation network input.} As mentioned in ~\cref{subsec:segmentation}, the input of the segmentation network is the element-wise product between the normalized feature maps of S-T networks as defined by \cref{eq:defX}. To prove the rationality of this setting, we build two distinct feature combinations as input. The first is to directly concatenate the feature maps of S-T networks $F_{S_k}$ and $F_{T_k}$ as the input of the segmentation network, which preserves the information of the S-T networks more effectively. The second is to compute the cosine distance of the S-T networks' feature maps using \cref{eq:defD}, which utilizes more prior information when we train the student network by optimizing the cosine distance. We show the results in \cref{tab:inputablation}. Both approaches result in suboptimal performance, indicating that $\hat{X}$ is a suitable choice as the input to balance the information and prior.

\begin{table}[htbp]
  \centering
  \begin{threeparttable}
  \setlength{\tabcolsep}{1.5mm}{
  \begin{tabular}{ccccccc}
    \toprule
    Exp. & den & ed & seg & img (AUC) & pix (AP) & ins (IAP) \\
    \midrule
    1&&&&94.8&52.9&55.8 \\
    2&$\checkmark$&&&93.4&49.6&53.9 \\
    3&&$\checkmark$&&95.4&53.3&57.7 \\
    4&&&$\checkmark$&97.3&70.1&71.8 \\
    5&$\checkmark$&$\checkmark$&&94.5&54.0&58.5 \\
    6&$\checkmark$&&$\checkmark$&97.3&70.9&72.3 \\
    7&&$\checkmark$&$\checkmark$&97.7&69.7&71.2 \\
    8&$\checkmark$&$\checkmark$&$\checkmark$&\pmb{98.6}&\pmb{75.8}&\pmb{76.4}\\
    \bottomrule
  \end{tabular}}
  \end{threeparttable}

  \caption{Ablation studies on our main designs: denoising training (den), the encoder-decoder architecture of student network (ed), and segmentation network (seg). AUC, AP, and IAP (\%) are used to evaluate image-level, pixel-level, and instance-level detection, respectively. Exp. 1 uses the same architecture of ~\cite{wang2021student}, but different training settings to align with Exp. 2\(\sim \)8.}
  \label{tab:mainablation}

\end{table}

\begin{table}[htbp]
  \centering
  \begin{threeparttable}
  \setlength{\tabcolsep}{1mm}{
  \begin{tabular}{lccc}
    \toprule
    & img (AUC) & pix (AP) & ins (IAP) \\
    \midrule
    % w/o data augmentation & 97.7 & 73.7 & 72.7 \\
    w/o L1 loss & 97.9 & 72.2 & 74.4 \\
    % concatenated-ST input & 98.0 & 72.2 & 72.6 \\
    % cosine-distance input & 98.5 & 72.0 & 74.5 \\
    w/ L1 loss & \pmb{98.6}&\pmb{75.8}&\pmb{76.4}\\
    \bottomrule
  \end{tabular}}
  \end{threeparttable}

  \caption{Ablation studies on the segmentation loss: AUC, AP, and IAP (\%) are used to evaluate image-level, pixel-level, and instance-level detection, respectively.}
  \label{tab:otherablation}

\end{table}

\begin{table}[htbp]
  \centering
  \begin{threeparttable}
  \setlength{\tabcolsep}{1mm}{
  \begin{tabular}{lccc}
    \toprule
    & img (AUC) & pix (AP) & ins (IAP) \\
    \midrule
    concatenated-ST input & 98.0 & 72.2 & 72.6 \\
    cosine-distance input & 98.5 & 72.0 & 74.5 \\
    DeSTSeg & \pmb{98.6}&\pmb{75.8}&\pmb{76.4}\\
    \bottomrule
  \end{tabular}}
  \end{threeparttable}

  \caption{Ablation studies on the input of segmentation network: AUC, AP, and IAP (\%) are used to evaluate image-level, pixel-level, and instance-level detection, respectively.}
  \label{tab:inputablation}

\end{table}

\section{Conclusion}
\label{sec:conclusion}

We propose the DeSTSeg, a segmentation-guided denoising student-teacher framework for the anomaly detection task. The denoising student-teacher network is adopted to enable the S-T network to generate discriminative features in anomalous regions. The segmentation network is built to fuse the S-T network features adaptively. Experiments on the surface anomaly detection benchmark show that all of our proposed components considerably boost performance. Our results outperform the previous state-of-the-art by 0.1\% AUC for image-level anomaly detection, 5.6\% AP for pixel-level anomaly localization, and 4.9\% IAP for instance-level anomaly detection. 

{\small
\bibliographystyle{ieee_fullname}
\bibliography{egbib}
}

\end{document}